# Computer sciences and synthesis: retrospective and perspective

Vladislav P. Dorofeev[1,a]

Petro P. Trokhimchuck[2,b]

[1] *Scientific Research Institute of System Analysis of Russian Academy of Sciences, Moscow, Russia*

*Phone: +7-9265622462*

[2]*Anatoliy Svidznskyi Department of the Theoretical and Computer Physics,*

*Lesya Ukrayinka Volyn National University, Voly av., 13, Lutsk, Ukraine, 43 025.*

*Phone: +38-0974421766*

[a]vladislav.dorofeev@gmail.com

[b]Trokhimchuck.Petro@vnu.edu.ua; trope1650@gmail.com

***Abstract.*** *The problem of synthesis in computer sciences, including cybernetics, artificial intelligence and system analysis, is analyzed. Main methods of realization this problem are discussed. Ways of search universal method of creation universal synthetic science are represented. As example of such universal method polymetric analysis is given. Perspective of further development of this research, including application polymetric method for the resolution main problems of computer sciences, is analyzed too.*

I. **Introduction**

The problem of synthesis in modern computer sciences is connected with the problem of creation the universal system of knowledge and has long history [1 – 17]. Creation the concept of modern science was beginning from Pythagor, Euclides, Plato and Aristotle. Later Pythagorean line (the synthesis) was developed by Descartes. He begin to realize the formalization of thesis "Science is science as science as it is mathematics" or Pythagorean phrase "Numbers are ruling of the world" [3]. His method was optimizing by I. Newton (four rules of conclusions in physics) [1]. This method is base in the creation of modern science [3]. But Newtonian rules are rules of selection and creation of concrete science. Development of modern science shows the necessity of creation more universal sciences [3, 4, 13, 14]. Leubniz's search and intentions to create a universal calculus led to the creation of mathematical logic. The greatest success in this had the work of B. Russell [10] and S. Klini [17]. B. Russell created the theory of logical types, which made it possible to inductively expand the application of logic for the analysis of more complex systems. S. Klini applied this in computer science [17]. For the tasks of artificial intelligence, this direction was developed by N. Nicholls [19 – 21].

Cybernetics as a science appeared in connection with the development of computer technology and with the need to process and operate with large amounts of information. According to N. Wiener cybernetics is science about control and communication in the animal and the machine [16, 22].

But, according to [3] cybernetics is synthesis of many sciences (mathematics, physics, biology, psychology and other).

Therefore the main problem of our research is ascertainment of question about possible application of PA for the resolution the problems of cybernetics, including general problems (S. Beer centurial problem, problem of complexity) and particular (matrix calculations, arrays sorting, pattern recognition).

Artificial intelligence (AI) is intelligence demonstrated by machines, as opposed to natural intelligence displayed by animals including humans. Leading AI textbooks define the field as the study of "intelligent agents": any system that perceives its environment and takes actions that maximize its chance of achieving its goals.[a] Some popular accounts use the term "artificial intelligence" to describe machines that mimic "cognitive" functions that humans associate with the human mind, such as "learning" and "problem solving", however, this definition is rejected by major AI researchers. It may be represented as chapter of cybernetics, but F. George is selected it as independent science.

Main ways of synthesis in artificial intelligence (logic, engineering and Moiseev type) are analyzed too. First two represented by S. Russel and P. Norwig [23], N. Nillson [19 – 21] and N. Kasabov [24] concepts, and have more engineering nature. Moiseev concept of hybrid super intelligence has more deep anthropic nature [25, 26]. This concept is connected with influence human activity on environment. Therefore, in this concept we have three principles: two ecological (Legasov and Efremov) and one system cybernetic (Moiseev) [26]. This concept was used to the resolution system aspects of COVID-19 [27]. These methods have inductive nature basically.

But cybernetics is synthetic science [16, 22]. Therefore, we must been create universal synthetic deductive theory. This theory is Polymetric Analysis.

More fully realization of this idea is Polymetric Analysis (PA): theory of variable measure or theory of systems with variable hierarchy. We can say that this system was created for the formalization L. Hall phrase "All, which come from head, is intelligent" [1, 3].

PA includes the procedure of measurement in finish result of measurement [13, 14]. Basic and derivative measurements are connected with quantitative and qualitative mathematical transformations. First is corresponded the procedure of measurement (arithmetic) the observed quantity, second – the analysis of dimensions (dimensional analysis).

Polymetric analysis is based on idea of triple minimum (particularly scientific, methodical and mathematical). Main principles of PA are criteria of reciprocity and simplicity. The first criterion is the principle of assembling the elements of the corresponding construct into a single system. Second criterion is principle optimality (simplicity complexity) of this assembling.

One of main component of this method, hybrid theory of systems (theory systems with variable hierarchy) show that only ten minimal types system of formalization the knowledge are existed [13, 14].

Therefore methods of PA as universal theory of optimal formalized synthesis may be used for the resolutions the main cybernetical problems. Structure of PA may be represented as more deep formalization the neuronets [1].

Therefore PA as universal system formalization of knowledge may be used for the resolution the basic problems of natural and artificial intelligence too [3, 13, 14].

The bonds of Polymetric Analysis and computer sciences are shown. The using of polymetric method for the resolution the problems of cybernetics and artificial intelligence is discussed. Perspective of using synthetic methods for the development of computer sciences is analyzed too.

## II. Cybernetics as synthetic science

Cybernetics (from the Greek κυβερνητική "governance," κυβερνώ "to steer, navigate or govern," κυβερνη "an administrative unit; an object of governance containing people") is the science of general regularities of control and information transmission processes in different systems, whether machines, animals or society [6, 16, 22].

Cybernetics studies the concepts of control and communication in living organisms, machines and organizations including self-organization. It focuses on how a (digital, mechanical or biological) system processes information, responds to it and changes or being changed for better functioning (including control and communication).

Cybernetics is an interdisciplinary science [4 – 7, 16, 22, 28 – 35]. It originated "at the junction" of mathematics, logic, semiotics, physiology, biology and sociology. Among its inherent features, we mention analysis and revelation of general principles and approaches in scientific cognition. Control theory, communication theory, operations research and others represent most weighty theories within cybernetics [22].

In ancient Greece, the term "cybernetics" denoted the art of a municipal governor (e.g., in Plato's Laws) [22].

H. Ampere (1834) related cybernetics to political sciences: he defined cybernetics ("the science of civil government") as a science of current policy and practical governance in a state or society) [6, 22].

B. Trentowsky (1843) viewed cybernetics as "the art of how to govern a nation" [14, 22].

In its "Tektology" (1925), A. Bogdanov examined common organizational principles for all types of systems. In fact, he anticipated many results of N. Wiener and L. von Bertalanffy [6], as the both were not familiar with Bogdanov's works [6].

The modern (and classical!) interpretation of the term "cybernetics" as "the scientific study of control and communication in the animal and the machine" was pioneered by Norbert Wiener in 1948, see the monograph [16]. Two years later, Wiener also added society as the third object of cybernetics [22]. Among other classics, we mention William Ashby [22] (1956) and Stafford Beer [22] (1959), who made their emphasis on the biological and "economic" aspects of cybernetics, respectively.

The latter case covers partial "intersection" of these results (see Fig. 1 – figuratively speaking, the central rode of the "umbrella"), i.e., usage of common results for all component sciences. Furthermore, we will adhere to this approach over and over again for discrimination between the corresponding umbrella brand and the common results of all component sciences in the context of different categories such as interdisciplinarity, systems analysis, organization theory, etc.

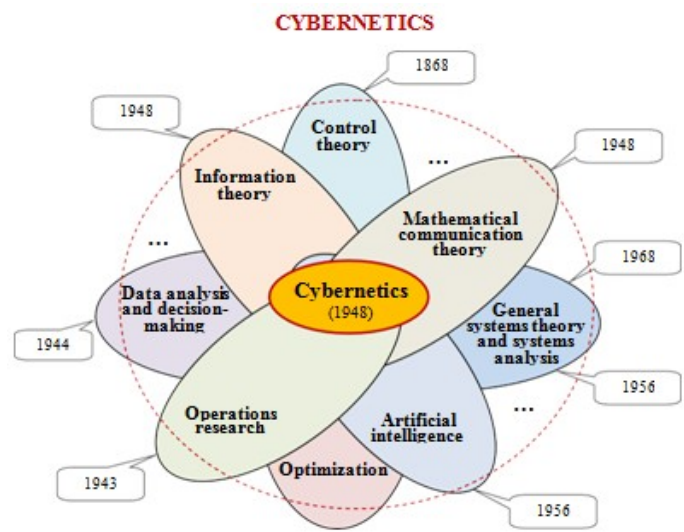

**Fig. 1.** The composition and structure of cybernetics [22].

Cybernetics today (disciplines included in cybernetics in the descending order of their "grades" of membership, see Fig. 1, with year of birth if available) [22]:

– control theory (1868–the papers published by J. Maxwell and I. Vyshnegradsky);

– mathematical theory of communication and information (1948 – C. Shannon's works);

– general systems theory, systems engineering and systems analysis;

– optimization (including linear and nonlinear programming; dynamic programming; optimal control; fuzzy optimization; discrete optimization, genetic algorithms, and so on);

– operations research (graph theory, game theory and statistical decisions, etc.);

– artificial intelligence (1956–The Dartmouth Summer Research Project on Artificial Intelligence);

– data analysis and decision-making;

– robotics

and others (purely mathematical and applied sciences and scientific directions, in an arbitrary order) including systems engineering, recognition, artificial neural networks and neural computers, ergatic systems, fuzzy systems (rough sets, grey systems, etc), mathematical logic, identification theory, algorithm theory, scheduling theory and queuing theory, mathematical linguistics, programming theory, synergetics and all similar sciences [22].

According to [4] cybernetics is synthesis of many sciences (mathematics, physics, biology, psychology and other Fig. 2.

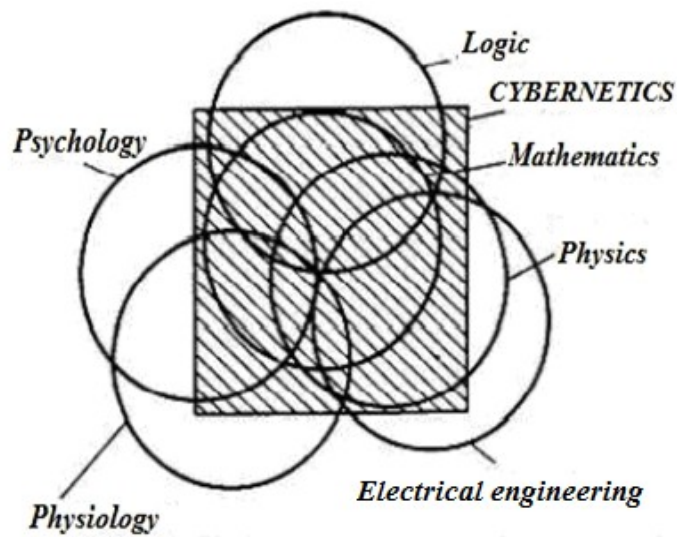

**Fig. 2.** A diagram that roughly illustrates the areas of intersection of the main disciplines that feed cybernetics [4].

Really this synthesis is more widely. For specific systems, this synthesis is quite general and therefore, as a rule, it is detailed. Therefore, in our times the new particular synthetic science, which are based on cybernetics are created. These particular synthetic sciences were called as biological cybernetics, economical cybernetics, physical cybernetics etc [4, 33].

There is one further argument which we will consider and that is that a computer, or any other artificially constructed system, only does what it is made to do by its programmer. This is a view that was held by Lady Lovelace [4]. It is a popular fallacy that computers can only do what the programmers make the computers do. The fallacy arises from various considerations. One is the failure to remember that human beings only do what they are programmed to do, although they are programmed by various different features of the environment, including parents, teachers, etc. and are adaptable and change according to changing circumstances. Now in this sense it is perfectly true to say that computers can only do what they are programmed to do, but they can certainly be given exactly the same flexibility as humans. In other words, various people can program them and they can be made adaptive so that they change and function in changing circumstances [4, 5].

The problem of not being able to do anything really new also relates in some measure to Lady Lovelace's objection which was really dependent upon the idea that the computer, or any other "machine" which was manufactured by human beings, could do no more than the programmer programmed it to do [4].

Schema of Fig. 1 shows the composition and structure cybernetics in historical way of its development. We see that artificial intelligence is the son and daughter of cybernetics, but it development in last years allow to select and represent the synthesis in this engineering science as separate paragraph.

### III. Artificial intelligence as synthetic science

Less than a decade after breaking the Nazi encryption machine Enigma and helping the Allied Forces win World War II, mathematician Alan Turing changed history a second time with a simple question: "Can machines think?" [4].

Turing's paper "Computing Machinery and Intelligence" (1950), and its subsequent Turing Test, established the fundamental goal and vision of artificial intelligence [4].

At its core, artificial intelligence (AI) is the branch of computer science that aims to answer Turing's question in the affirmative. It is the endeavor to replicate or simulate human intelligence in machines.

The expansive goal of artificial intelligence has given rise to many questions and debates. So much so, that no singular definition of the field is universally accepted.

The major limitation in defining AI as simply "building machines that are intelligent" is that it doesn't actually explain *what artificial intelligence is? What makes a machine intelligent?* AI is an interdisciplinary science with multiple approaches, but advancements in machine learning and deep learning are creating a paradigm shift in virtually every sector of the tech industry [4, 23].

According to N. Moiseev "Anyway, the term "artificial intelligence" established in the scientific literature, and with this it follows to be considered. However, it is important to clearly stipulate pragmatic, applied meaning of this term. So we agree to associate its use only with modern processing technology and use information" [17].

In their groundbreaking textbook *Artificial Intelligence: A Modern Approach*, authors Stuart Russell and Peter Norvig approach the question by unifying their work around the theme of intelligent agents in machines. With this in mind, AI is "the study of agents that receive percepts from the environment and perform actions" [23]

Norvig and Russell go on to explore four different approaches that have historically defined the field of AI [23]:

1. Thinking humanly

2. Thinking rationally

3. Acting humanly

4. Acting rationally

The first two ideas concern thought processes and reasoning, while the others deal with behavior. Norvig and Russell focus particularly on rational agents that act to achieve the best outcome, noting "all the skills needed for the Turing Test also allow an agent to act rationally" [23].

Winston, the Ford professor of artificial intelligence and computer science at MIT, defines AI as "algorithms enabled by constraints, exposed by representations that support models targeted at loops that tie thinking, perception and action together" [23].

While these definitions may seem abstract to the average person, they help focus the field as an area of computer science and provide a blueprint for infusing machines and programs with machine learning and other subsets of artificial intelligence.

***The Four Types of Artificial Intelligence are existed***:

1. *Reactive Machines.*

A reactive machine follows the most basic of AI principles and, as its name implies, is capable of only using its intelligence to perceive and react to the world in front of it. A reactive machine cannot store a memory and as a result cannot rely on past experiences to inform decision making in real-time.

Perceiving the world directly means that reactive machines are designed to complete only a limited number of specialized duties. Intentionally narrowing a reactive machine's worldview is not any sort of cost-cutting measure, however, and instead means that this type of AI will be more trustworthy and reliable — it will react the same way to the same stimuli every time.

A famous example of a reactive machine is Deep Blue, which was designed by IBM in the 1990's as a chess-playing supercomputer and defeated international grandmaster Gary Kasparov in a game. Deep Blue was only capable of identifying the pieces on a chess board and knowing how each moves based on the rules of chess, acknowledging each piece's present position, and determining what the most logical move would be at that moment. The computer was not pursuing future potential moves by its opponent or trying to put its own pieces in better position. Every turn was viewed as its own reality, separate from any other movement that was made beforehand.

Another example of a game-playing reactive machine is Google's AlphaGo. AlphaGo is also incapable of evaluating future moves but relies on its own neural network to evaluate developments of the present game, giving it an edge over Deep Blue in a more complex game. AlphaGo also bested world-class competitors of the game, defeating champion Go player Lee Sedol in 2016.

Though limited in scope and not easily altered, reactive machine artificial intelligence can attain a level of complexity, and offers reliability when created to fulfill repeatable tasks.

2. *Limited Memory.*

Limited memory artificial intelligence has the ability to store previous data and predictions when gathering information and weighing potential decisions — essentially looking into the past for clues on what may come next. Limited memory artificial intelligence is more complex and presents greater possibilities than reactive machines.

Limited memory AI is created when a team continuously trains a model in how to analyze and utilize new data or an AI environment is built so models can be automatically trained and renewed. When utilizing limited memory AI in machine learning, six steps must be followed: Training data must be created, the machine learning model must be created, the model must be able to make predictions, the model must be able to receive human or environmental feedback, that feedback must be stored as data, and these these steps must be reiterated as a cycle.

There are three major machine learning models that utilize limited memory artificial intelligence:

• Reinforcement learning, which learns to make better predictions through repeated trial-and-error.

• Long Short Term Memory (LSTM), which utilizes past data to help predict the next item in a sequence. LTSMs view more recent information as most important when making predictions and discounts data from further in the past, though still utilizing it to form conclusions

• Evolutionary Generative Adversarial Networks (E-GAN), which evolves over time, growing to explore slightly modified paths based off of previous experiences with every new decision. This model is constantly in pursuit of a better path and utilizes simulations and statistics, or chance, to predict outcomes throughout its evolutionary mutation cycle.

3. *Theory of Mind.*

Theory of Mind is just that — theoretical. We have not yet achieved the technological and scientific capabilities necessary to reach this next level of artificial intelligence.

The concept is based on the psychological premise of understanding that other living things have thoughts and emotions that affect the behavior of one's self. In terms of AI machines, this would mean that AI could comprehend how humans, animals and other machines feel and make decisions through self-reflection and determination, and then will utilize that information to make decisions of their own. Essentially, machines would have to be able to grasp and process the concept of "mind," the fluctuations of emotions in decision making and a litany of other psychological concepts in real time, creating a two-way relationship between people and artificial intelligence.

4. *Self-awareness.*

Once Theory of Mind can be established in artificial intelligence, sometime well into the future, the final step will be for AI to become self-aware. This kind of artificial intelligence possesses human-level consciousness and understands its own existence in the world, as well as the presence and emotional state of others. It would be able to understand what others may need based on not just what they communicate to them but how they communicate it.

Self-awareness in artificial intelligence relies both on human researchers understanding the premise of consciousness and then learning how to replicate that so it can be built into machines.

According N. Nillson [19 – 21] artificial intelligence, broadly (and somewhat circularly) defined, is concerned with intelligent behavior in artifacts. Intelligent behavior, in turn, involves perception, reasoning, learning, communicating, and acting in complex environments. Al has as one of its long-term goals the development of machines that can do these things as well as humans can, or possibly even better. Another goal of artificial intelligence is to understand whether it occurs in machines or in humans or other animals. Thus, artificial intelligence has both engineering and scientific goals. N. Nillson is focused main attention on the important concepts and ideas underlying the design of intelligent machines.

Main elements of N. Nillson synthesis are [19 – 21]:

1. Reactive machines, which are included: stimulus-response agents (perception and action, representing and implementing action functions); neural networks, machine evolution (evolutionary computation, genetic programming); state machines (representing the environment by feature vectors, Elman networks, iconic representations, blackboard systems); robot vision.

2. Search in state spaces: agents that plan (memory versus computation, state-space graphs, searching explicit state spaces, feature-based state spaces, graph notation); uniformed search; heuristic search; planning, acting and learning; alternative search formulations and applications; adversarial search.

3. Knowledge representation and reasoning: the propositional calculus; resolution in the propositional calculus; the predicate calculus; resolution in the predicate calculus; knowledge-based systems; representing commonsense knowledge; reasoning with uncertain information; learning and acting with Bayes nets.

4. Planning methods based on Logic: the situation calculus; planning.

5. Communication and integration; multiple agents; communications among agents; agent architectures.

N. Kasabov engineering synthesis of artificial intelligence including four areas [24]:

1. Foundations: evolving processes and their representation as data, information and knowledge; brain information processing, evolutionary computation; quantum inspired computation; molecular information processing; information theory; computational architecture.

2. ECOS and SNN methods: artificial neural networks (ANN) and evolving connectionist system (ECOS); spiking neural networks (SNN) methods; ANN and ECOS computational methods; SNN methods; evolving SNN (eSNN); brain inspired SNN (BI-SNN) and the design brain inspired – artificial intelligence; SNN, eSNN, BI-SNN parameter optimization with evolutionary computation (EC).

3. Applications: deep learning and deep knowledge from brain data; audio- and visual information processing; bioinformatics data modeling; SNN for neuroinformatics and personalized modeling; predictive modeling in ecology; predictive modeling in transport; predictive in environment.

4. Future directions: brain-computer interfaces with BI-SNN; affective computation; neuromorphic systems; new spike-time information theory for data compression; integrated quantum-neurogenetic-brain-inspired models; towards integrated human intelligence and artificial intelligence.

Now we represent the model of hybrid super intelligence proposed in [25, 26], which is based on N. Moiseev's idea of complex modeling of the noosphere [17] in order to predict the consequences of any external influences on it, including anthropogenic ones [26].

The proposed strong hybrid intelligence architecture is shown in Fig.3 [25, 26]. The explanation of schema of Fig. 3 is next [25]:

1. The real world is information about the problem area, collected using the sensors available to the system. The real world includes both real world objects and automated transactional information processing systems. For example, in the case of a pandemic, contact tracing systems, medical information systems with patient data, etc.

2. People – a team of specialists involved in solving a problem. The team may include subject matter experts, developers, information system operators, etc.

3. Artificial intelligence is an adaptable and developed system for the automated collection and processing of real-world information with an interface for communication with a group of experts, including in natural language.

4. Modeling system – a set of systems for modeling and forecasting the real world with the capabilities of scenario analysis of the consequences of impact on the real world.

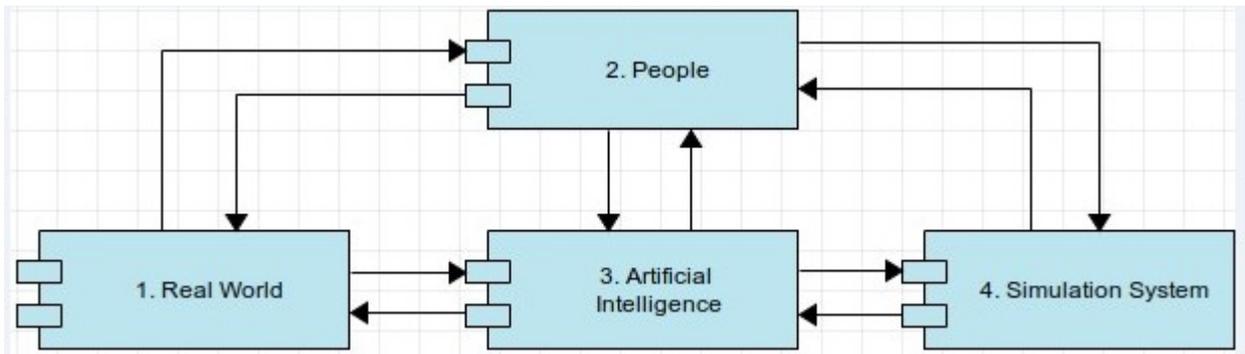

**Fig. 3**. Architecture the hybrid super intelligence of N. Moiseev type [25]

The goal of the system of strong hybrid intelligence is the most accurate forecasting of the development of the real world with the possibility of scenario analysis of the consequences of external influences on it.

In [27], this architecture was proposed to solve the COVID-19 problem. A simple experiment was carried out to expand the epidemiological SIRD-model with the parameter "The level of immunity of the population". A similar architecture was used on a larger scale in Canada [27]. As a real world, it used a contact monitoring system and medical information systems with patient data. Artificial intelligence tools were used to refine the parameters of the model system based on real-world data. The modeling system made it possible to carry out scenario analysis of the development of the situation. An international team of specialists ensured the development and use of the system.

In a sense, the proposed architecture can be considered an extension of the concept of "digital twins", developed in the world since 2002, to include the human factor. And at present, the main successes in the field of solving complex problems are associated with precisely such systems [25, 26].

The architecture of the proposed system itself assumes its transparency and controllability, since the consequences of the supposed impacts on the real world are checked on the modeling system. But taking into account the complexity of the real world and the limited possibilities of methods for its modeling, there are no complete guarantees of the safety of such a system. The following principles can be used to improve security.

1. Legasov's principle: a technostructure uncontrolled by society threatens the global security of mankind.

The first principle concerns the problem of control of the technostructure, which controls the system of strong artificial intelligence, by the society. It appeared on the basis of V. Legasov's notes on the circumstances of the Chernobyl accident [25]. In relation to the proposed system of strong hybrid intelligence, it means the following: transparency and availability of the information systems used.

2. Efremov's principle [25]: anthropogenic changes in the environment, the rate of which exceeds the physical, biological and social mechanisms of adaptation to them, carry the risks of destroying life.

The second principle is related to the ability of living systems to adapt to changes in the external environment, which the strong intelligence system plans to carry out. If the speed of adaptation is not enough, then it is better to refrain from making changes. It was formulated on the basis of an episode with a visit by earthlings to the planet Zirda from the book "The Andromeda Nebula" by I. Efremov [25]. Only poppies remained on the planet due to the underestimation of the danger by its inhabitants of the use of low-power ionizing radiation, contributing to a high rate of uncontrolled mutations.

3. Moiseev's principle [25]: the complexity of the models of the world should be comparable to the complexity of the problems they are designed to solve.

The third principle is designed to prevent the use of primitive models to solve complex problems. It was formulated on the basis of the works of N. Moiseev [7, 17]. Despite some advances, current modeling capabilities of real-world systems are limited in their ability to accurately reproduce the real world. And often researchers forget about this and go far from the real world in their theoretical constructions.

As we see, a tendency of differentiation is characterized AI too. Represented concepts are show this picture.

As we see, main synthetic concepts of artificial intelligence (Nillson, Russel and Norwig, Kasabov and Moiseev) have more inductive and inductive with elements deduction nature and haven't general value for all computer sciences. It may be used for the concrete problems ov computer sciences and depending from level of development of modern electronics and information technologies. Therefore, we must search method, which may be represented as universal system formalization of knowledge, including computer sciences [1 – 17, 36 – 38]. This method should be deductive in nature and include not only the rules of logical formalization, like the Leubniz-Russell-Klini-Nillson approach [13, 14], but also based on the nature of mathematics: analysis, synthesis and formalization of any field of knowledge. This method can be polymetric analysis.

### IV. Polymetric analysis as operational concept of computing science

Polymetric analysis (PA) was created as alternative optimal concept to logical, formal and constructive conceptions of modern mathematics and theory of information [13, 14]. This concept is based on the idea of triple minimum: mathematical, methodological and concrete scientific.

However, one of the main tasks of polymetric analysis is the problem of simplicity-complexity that arises when creating or solving a particular problem or science. It must be open system [13, 14].

In methodological sense, PA is the synthesis of Archimedes thesis: "Give me a fulcrum and I will move the world", and S. Beer idea about what complexity is a problem in cybernetics century, in one system. And as cybernetics is a synthetic science, the problem should be transferred and for all of modern science. Basic elements of this theory and their bonds with other science are represented in Fig. 4 [13, 14].

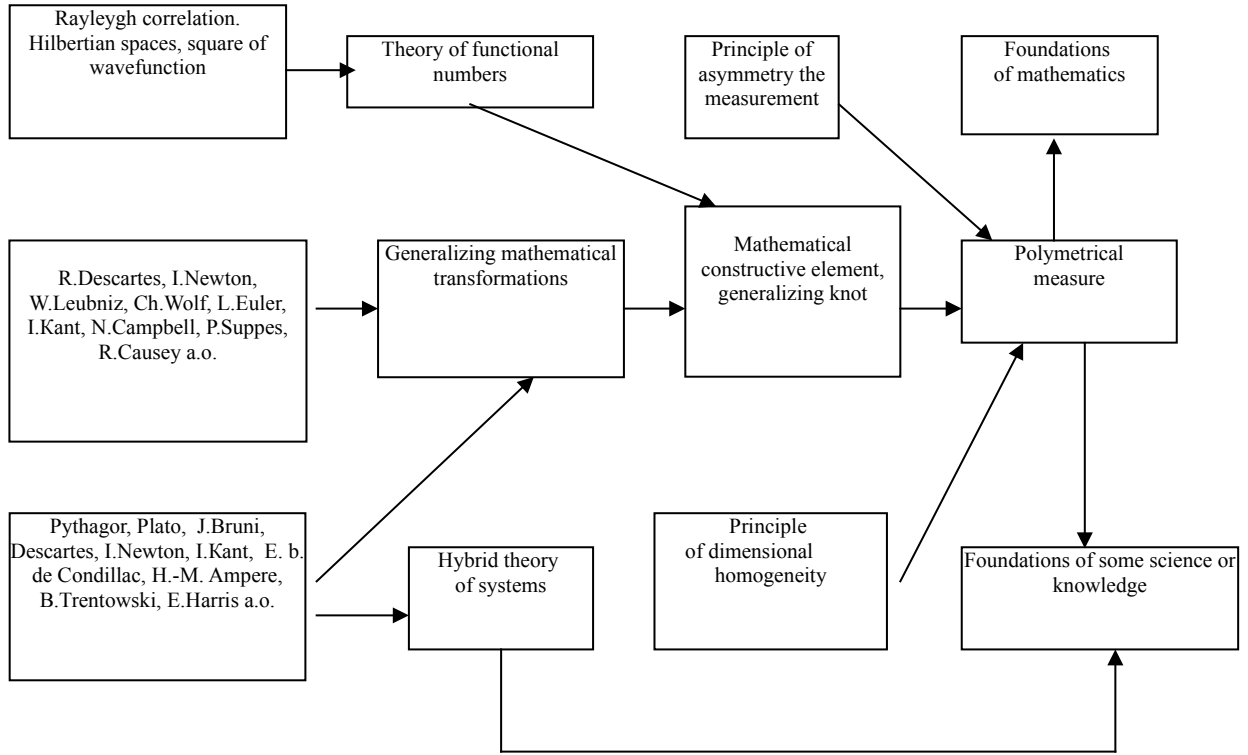

**Fig. 4.** Schema of polymetric method and its place in modern science [13, 14].

The polymetric analysis may be represented as universal theory of synthesis in Descartian sense. For resolution of this problem we must select basic notions and concepts, which are corresponded to optimal basic three directions of Fig. 4. The universal simple value is unit symbol, but this symbol must be connected with calculation. Therefore it must be number. For the compositions of these symbols (numbers) in one system we must use system control and operations (mathematical operations or transformations). After this procedure we received the proper measure, which is corresponding system of knowledge and science.

Therefore the basic axiomatic of the polymetric analysis is was selected in the next form [3]. This form is corresponded to schema of Fig, 4.

*Definition* 1 . **Mathematical construction** is called set all possible elements, operations and transformations for resolution corresponding problem. The basic functional elements of this construction are called constructive elements.

*Definition* 2 . The mathematical constructive elements $N_{x_{ij}}$ are called **the functional parameters**

$$N_{x_{ij}} = x_i \cdot \bar{x}_j, \qquad (1)$$

where $x_i$, $\overline{x_j}$ – the straight and opposite parameters, respectively; · – respective mathematical operation.

*Definition* 3. The mathematical constructive elements $N_{\varphi_{ij}}$ are called the **functional numbers**

$$N_{\varphi_{ij}} = \varphi_i \circ \overline{\varphi}_j. \qquad (2)$$

Where $\varphi_i(x_1,...x_n, \overline{x_1}...\overline{x_m},...N_{x_{ij}},...)$, $\overline{\varphi}_j(x_1,...x_n, \overline{x_1}...\overline{x_m},...N_{x_{ij}},...)$ are the straight and opposite functions, respectively; $\circ$ – respective mathematical operation.

*Remark* 1. Functions $\varphi_I$, $\overline{\varphi}_j$ may be have different nature: mathematical, linguistic and other.

*Definition* 4. The mathematical constructive elements $N^d_{x_{ij}}$ are called the **diagonal functional parameters**

$$N^d_{x_{ij}} = \delta_{ij} N_{x_{ij}}. \qquad (3)$$

Where $\delta_{ij}$ is Cronecker symbol.

*Definition* 5. The mathematical constructive elements $N^d_{\varphi_{ij}}$ are called the **diagonal functional numbers**

$$N^d_{\varphi_{ij}} = \delta_{ij} N_{\varphi_{ij}}. \qquad (4)$$

*Example* 1. If $x_i = x^i$, $\overline{x}_j = x^{-j}$ and $\max\{i\} = \max\{j\} = m$, then $\{N^d_{\varphi_{ij}}\}$ is diagonal single matrix. Another example may be the orthogonal eigenfunctions of the Hermitian operator.

*Remark* 2. These two examples illustrate why quantities (1) – (4) are called the parameters and numbers. Practically it is the simple formalization the measurable procedure in its logic and chaos parts [13, 14]. The straight functions correspond the "straight" observation and measurement and opposite functions correspond the "opposite" observation and measurement. This procedure is included in quantum mechanics the Hilbert's spaces and Hermitian operators.

The theory of generalizing mathematical transformations is created for works on functional numbers.

*Definition* 6. **Qualitative transformations** on functional numbers $N_{\varphi_{ij}}$ (straight $A_i$ and opposite $\overline{A}_j$) are called the next transformations. The straight qualitative transformations are reduced the dimension $N_{\varphi_{ij}}$ on *i* units for straight parameters, and the opposite qualitative transformations are reduced the dimension $N_{\varphi_{ij}}$ on *j* units for opposite parameters.

*Definition* 7. **Quantitative (calculative) transformations** on functional numbers $N_{\varphi_{ij}}$ (straight $O_k$ and opposite $\overline{O}_p$) are called the next transformations. The straight calculative transformations are reduced $N_{\varphi_{ij}}$ or corresponding mathematical constructive element on *k* units its measure. The opposite quantitative transformations are increased $N_{\varphi_{ij}}$ or corresponding mathematical constructive element on $l$ units its measure, i.e.

$$O_k \overline{O}_l N_{\varphi_{ij}} = N_{\varphi_{ij}} - k \oplus l. \qquad (5)$$

*Definition* 8. **Left and right transformations** are called transformations which act on left or right part of functional number respectively.

*Definition* 9. The maximal possible number corresponding transformations is called **the rang of this transformation**

$$rang(A_i \overline{A}_j N_{\varphi_{ij}}) = \max(i,j); \qquad (6)$$

$$rang(O_k \overline{O}_p N_{\varphi_{ij}}) = \max(k,p). \qquad (7)$$

*Remark* 3. The indexes *i,j, k,p* are called **the steps of the corresponding transformations.**

Only 15 minimal types of of generalizing mathematical transformations are existed [13, 14].

Basic elements of PA is the generalizing mathematical elements or its various presentations – informative knots. Generalizing mathematical element is the composition of functional numbers (generalizing quadratic forms, including complex numbers and functions) and generalizing mathematical transformations, which are acted on these functional numbers in whole or its elements. Roughly speaking these elements are elements of functional matrixes.

This element $^{stqo}_{nmab}M_{ijkp}$ may be represented in next form

$$^{stqo}_{nmab}M_{ijkp} = A_i \overline{A}_j O_k \overline{O}_p A_s^r \overline{A}_t^r O_q^r \overline{O}_o^r A_n^l \overline{A}_m^l O_a^l \overline{O}_b^l N_{\varphi_{ij}}. \qquad (8)$$

Where $N_{\varphi_{ij}}$ – functional number; $O_k, O_q^r, O_a^l, \overline{O}_p, \overline{O}_o^r, \overline{O}_b^l$; $A_i, A_s^r, A_n^l, \overline{A}_j, \overline{A}_t^r, \overline{A}_m^l$ are quantitative and qualitative transformations, straight and inverse (with tilde), (r) – right and (l) – left.

Polyfunctional matrix, which is constructed on elements (8) is called informative lattice. For this case generalizing mathematical element was called knot of informative lattice [3]. Informative lattice is basic set of theory of informative calculations. This theory was constructed analogously to the analytical mechanics.

Basic elements of this theory are:

1. Informative computability *C* is number of possible mathematical operations, which are required for the resolution of proper problem.

2. Technical informative computability $C_t = C \Sigma t_i$, where $t_i$ – realization time of proper computation.

3. Generalizing technical informative computability $C_{t0} = k_{ac} C_t$, where $k_{ac}$ – a coefficient of algorithmic complexity.

Basic principle of this theory is **the principle of optimal informative calculations**: any algebraic, including constructive, informative problem has optimal resolution for minimum informative computability $C$, technical informative computability $C_t$ or generalizing technical informative computability $C_{to}$.

The principle of optimal informative calculations is analogous to action and entropy (second law of thermodynamics) principles in physics.

The principle of optimal informative calculation is more general than **negentropic principle the theory of the information** [28] and **Shennon theorem** [29]. This principle is law of the open systems or systems with variable hierarchy. The negenthropic principle and Shennon theorem are the principles of systems with constant hierarchy.

Idea of this principle of optimal informative calculation may be explained on the basis de Broglie formula [39]

$$S_a/\hbar = S_e/k_B \quad (9)$$

(equivalence of quantity of ordered and disorder information). Where $S_a$ – action, $\hbar$ – Planck constant, $S_e$ – entropy and $k_B$ – Boltzman constant. Therefore we can go from dimensional quantities (action and entropy) to undimensional quantity – number of proper quanta or after generalization to number of mathematical operations. Thus, theory of informative calculations may be represented as numerical generalization of classical theory of information.

For classification the computations on informative lattices hybrid theory of systems was created [13, 14]. This theory allow to analyze proper system with point of view of its complexity,

The basic principles of hybrid theory of systems are next: 1) **the criterion of reciprocity;** 2) **the criterion of simplicity**.

The criterion of reciprocity is the principle of the creation the corresponding mathematical constructive system (informative lattice). The criterion of simplicity is the principle the optimization of this creation.

The basic axiomatic of hybrid theory of systems (HTS) is represented below.

*Definition* 10. The set of functional numbers and generalizing transformations together with principles reciprocity and simplicity (informative lattice) is called **the hybrid theory of systems** (in more narrow sense the criterion of the reciprocity and principle of optimal informative calculations).

*Criterion of the reciprocity* for corresponding systems is signed the conservation in these systems the next categories:

1) the completeness;

2) the equilibrium;

3) the equality of the number epistemological equivalent known and unknown knotions.

*Criterion of the simplicity* for corresponding systems is signed the conservation in these systems the next categories:

1) the completeness;

2) the equilibrium;

3) the principle of the optimal informative calculative calculations.

Criterion of reciprocity is the principle of creation of proper informative lattice. Basic elements of principle reciprocity are various nuances of completeness. Criterion of the simplicity is the principle of the optimality of this creation.

For more full formalization the all famous regions of knowledge and science the **parameter of connectedness** $\sigma_t$ was introduced. This parameter is meant the number of different bounds the one element of mathematical construction with other elements of this construction. For example, in classic mathematics $\sigma_t = 1$, in linguistics and semiotics $\sigma_t > 1$. The parameter of connectedness is the basic element for synthesis in one system of formalization the all famous regions of knowledge and science. It is one of the basic elements for creation the theory of functional logical automata too.

For classification of systems of calculation hybrid theory of systems was created. This theory is based on two criterions: criterion of reciprocity – principle of creation of proper formal system, and criterion of simplicity – principle of optimality of this creation. For "inner" bond of two elements of informative lattice a parameter of connectedness $\sigma_t$ was introduced. Principle of optimal informative calculation is included in criterion of simplicity.

At help these criteria of reciprocity and simplicity and parameter of connectedness the basic famous parts of knowledge and science may be represent as next 10 types of hybrid systems [13, 14]:

1. The system with conservation all positions the criteria of reciprocity and simplicity for all elements of mathematical construction ( $N_{\varphi_{ij}}$ and transformations) is called the *simple system*.

2. The system with conservation the criterion of simplicity only for $N_{\varphi_{ij}}$ is called the *parametric simple system*.

*Remark* 4. Further in this classification reminder of criteria of reciprocity and simplicity is absented. It mean that these criteria for next types of hybrid systems are true.

3. The system with conservation the criterion of simplicity only for general mathematical transformations is called *functional simple system*.

4. The system with nonconservation the principle of optimal informative calculation and with $\sigma_t = 1$ is called the *semisimple system*.

5. The system with nonconservation the principle of optimal informative calculation only for $N_{\varphi_{ij}}$ and with $\sigma_t = 1$ is called the *parametric semisimple system*.

6. The system with nonconservation the principle of optimal informative calculation only for general mathematical transformations and with $\sigma_t = 1$ is called the *functional semisimple system*.

7. The system with nonconservation the principle of optimal informative calculation and with $\sigma_t \neq 1$ is called *complicated system.*

8. The system with nonconservation the principle of optimal informative calculation only for $N_{\varphi_{ij}}$ is called *parametric complicated system.*

9. The system with nonconservation the principle of optimal informative calculation only for general mathematical transformations and with $\sigma_t \neq 1$ is called *functional complicated system*.

10. The system with nonconservation the criteriums of reciprocity and simplicity and with $\sigma_t \neq 1$ is called *absolute complicated system*.

With taking into account 15 basic types of generalized mathematical transformations we have 150 types of hybrid systems; practically 150 types of the formalization and modeling of knowledge and science.

Only first six types of hybrid systems may be considered as mathematical, last four types are not mathematically. Therefore HTS may be describing all possible system of knowledge. Problem of verbal and nonverbal systems of knowledge is controlled with help of types the mathematical transformations and parameter connectedness [13, 14].

This theory has finite number of types the knowledge formalization systems. And in general, this theory is the theory of open systems with a changeable hierarchy.

Therefore, HTS with its operational nature may be used for all knowledge and culture, including cybernetics and artificial intelligence.

## V. Polymetric analysis and computer sciences

We can analyze PA and computer sciences with point of conditions, which are formulated for the general theories (theories of everything) [1, 3]:

1. It must be open theory or theory with variable hierarchy.
2. This theory must be having minimal number of principles.
3. It must based on nature of mathematics (analysis, synthesis and formalization all possible knowledge).
4. We must create sign structure, which unite verbal and nonverbal knowledge (mathematical and other) in one system.
5. We must have system, which is expert system of existing system of knowledge and may be use for the creation new systems of knowledge.
6. Principle of continuity must be true for all science.

These conditions must be used for the creation any dynamic science, which can be presented as open system.

These conditions were formulated on the basis of polymetric analysis. But other theories of everything may be creating according to these six conditions.

But main conditions for the computer systems must be: completeness, unambiguity, simplicity and possibilities to create corresponding system and measure and estimate proper set of information in this system []. But in real computer system we must include corresponding technologies, which which impose limitations on computing capabilities, especially performance. Further, we must transform all possible information in the form convenient for its processing by a computer processor, and then its adequate presentation for the user. But modern processors worked with various matrixes, therefore informative lattice of generalized constructive elements may be represented as mathematical generalization of computer processor.

The one of central problem of modern computing sciences is problem of information complexity [1, 2]. This problem was formulated in cybernetics by S. Beer (S. Beer centurial problem in cybernetics). This formulation is next [2]: "Apparently, the complexity becomes the problem of the century, just as the ability to process natural materials has been a problem of life and death for our forefathers. Our tool must be computers, and their efficiency should be provided by science, able to handle large and complex systems of probabilistic nature. This science may be cybernetics – the science of management processes and communication. The basic thesis of cybernetics can be set forth as follows: there are natural laws behavior of the large multibonds systems of any character submits that – biological, technical, special and economic."

In whole problem of complexity have two sides. First is pure computational. This problem is represented by two Smale's problems [14, 40, 41], which are not only problems of modern computing science but it are problems of modern mathematics. The complexity of networks is represented as problem of complexity in modern physics [1]. But neuronets were introduced in cybernetics: M. Minsky as perceptrons [1] and A. Ivakhnenko [1]. Therefore this problem is cybernetic too. In this case we have third Kolmogorov algorithmic concept [1] in theory of information. But according F. H. George [5] cybernetics is the synthesis of many sciences: physics, mathematics, biology, psychology, linguistics and other. Therefore we must add three Kolmogorov concepts in information theory by fourth system concept. This concept is second side of problem complexity in modern science. It has more cybernetic and computing science nature as physical [1]. Problem of complexity is the basic of other computing sciences, including artificial intelligence, and as "sons" and "daughters" of cybernetics too [1]. Therefore system concept of complexity is more necessary for cybernetics and computing science as for physics because its sciences have more synthetic nature as physics [1].

Roughly speaking the ways of search the resolutions of this problem may be formulated with help phrase of Ukrainian philosopher G. Skovoroda: "Thank You God, He made all the essentials simple and understandable, and all the necessary things difficult and incomprehensible" [14].

Therefore methods of PA as universal theory of optimal formalized synthesis may be used for the resolutions the main cybernetic problems. Therefore the main problem of our paper is ascertainment of question about possible application of PA for the resolution the problems of cybernetics, including general problems (S. Beer centurial problem, problem of complexity) and particular (matrix calculations, arrays sorting, pattern recognition).

Structure of PA may be represented as more deep formalization the neuronets too [1, 3].

According to F. George "The brain is universal computer" [4, 6]. Therefore PA as universal system formalization of knowledge may be used for the resolution the basic problems of natural and artificial intelligence too [1, 3].

Polymetric analysis fully satisfies these conditions, the represented cybernetics and artificial intelligence concepts and systems – partially.

Polymetrical Analysis is more general system as cybernetics and artificial intelligence of all represented types. It is operational system, which is included the procedure of measurement with help generalizing mathematical transformations. Generalized constructive element (8) is term of polyfunctional matrix. But computer processors are using matrix calculation [1]. Therefore PA may be represented as functional expansion of computer processor, which are include the procedures of collection and processing of information using generalized mathematical transformations and criteria of reciprocity and simplicity.

More concrete analysis of hybrid super intelligence of N. Moiseev type we are made [26].

Hybrid super intelligence of N. Moiseev type has more narrow scope. Therefore it is more anthropic system as PA [26]. Roughly speaking Legasov, Efremov principles are anthropic principles and only Moiseev principle is connected with procedure of system formalization.

Legasov principle is ecological principle and has more long history. These are the problems of chemical, nuclear and atmospheric pollution of the environment. These problems were raised by physicists themselves such as A. Einstein, N. Bohr, A. Sakharov, philosophers – B. Russell and other (Russel-Einstein manifesto, Pugwash Conferences on Science and World Affairs and Greenpeece) [42]. The same political movement of the greens arose. Based on this, the international independent non-governmental environmental organization Greenpeace was established in 1971 in Canada [15].

The organization's focus is on issues such as global climate change, deforestation from the tropics to the Arctic to the Antarctic, overfishing, commercial whaling, radiation hazards, renewable energy and resource conservation, hazardous chemical pollution, sustainable agriculture, economy, preservation of the Arctic.

Therefore Legasov and Efremov principles have ecological nature. Legasov principle is characterized the irreversible change in nature, Efremov principle is corresponded to the establishing the boundaries of this irreversibility.

From mathematical point of view the Legasov principle may be represented as boundary conditions on corresponding qualitative transformations, Efremov principle imposes an additional cyclic condition, which must be imposed on the corresponding mathematical construct. For the Efremov principle, a reversible reproducible feedback is important, which can be set both through inverse mathematical transformations and through their combinations.

The division of complexity into subject and mathematical is quite conventional. In polymetric analysis, it has a systematic form with an emphasis on calculation. The fact that the complexity of information processing needs to be carried out through calculations was pointed out by Kasti [32].

The Moiseev principle, the principle of correspondence between the complexity of the formalized (modeled) area and formalization (modeling) methods may be represented as equivalence three provisions of the principle of reciprocity, especially the first two provisions with the third. Really in computer science problem of complexity must reduce to problem of complexity the calculation in more general sense. For PA it must be more optimal principle with calculation point of view.

Other chapters of computer sciences, including cybernetics, artificial intelligence, computer arithmetics may be expand, represent and explain with help Polymetric Analysis.

So, theory of informative calculations and principle of optimal informative calculations allow resolving many computer problems of modern cybernetics in area of the obtained algorithms, matrix algebra and the problem of forming arrays. From a fundamental point of view, the principle of optimal informational calculations makes it possible to bring physical processes and information theory closer together. This is shown in the theory of information-physical structures.

From a systemic point of view, polymetric analysis is a synthetic optimal extension of those concepts and methods that played a decisive role in the formation and development of both modern science and other areas of knowledge and culture. It can be used both to determine the complexity of computations (theory of information calculations) and to determine the complexity of systems and in the choice of both an expert system and a new promising system for the corresponding synthesis (theory of hybrid systems) [13, 14].

The way of the resolution the problem of complexity in computer sciences must be connected with problem of calculations [13, 14]].

Roughly speaking modern computing science may be represented as renewal of Pythagorean system [14]. Three typed numbers (mathematical, sensitive and ideal) were in Plato's school [14]. With modern point of view mathematical numbers are corresponded to pure mathematics; sensitive numbers – applied mathematics; ideal numbers – other areas of knowledge [14]. According to this fact, we must include all three types of these numbers in one system, which must correspond to our six conditions for theories of everything (universal system of knowledge) [14]. But modern science is more science and knowledge is richer as in the time of the antiquity. But idea of triple minimum, which is basic for polymetric analysis, may be connected with Plato's numbers too.

In modern computer science Godel's numbers (number has system nature) [4, 37] may be represented as renewal of Plato concept in modern science too [14]. Each universal system of knowledge, including computer sciences must include these aspects for the creation.

In whole, the creation of effective computer science must be include general system principles, which are corresponded its universality and particular principles, models and theories, which are corresponded to modern level of information and optoelectronics and may be in nearest future biological technologies. In last case the idea of rapprochement of the living and inanimate worlds, including intelligence, can be more fully realized than at the same stage in the development of science and technology. To solve this problem, we must look for both new ways of both formalizing knowledge and their hardware representation and processing.

Thus in this work we represented the main peculiarities of modern state computer sciences on the examples of cybernetics and artificial intelligence and shown the perspectives of creation universal theory of everything, which are based on modern computer sciences.

## VI. Conclusions

1. The problems of synthesis in computer sciences are analyzed.
2. Short analysis cybernetics as synthetic science is represented.
3. Artificial intelligence and synthetic aspects of its development are observed.
4. Basic concepts of Polymetric Analysis as universal system of formalization the knowledge are analyzed.
5. Questions of using Polymetric Analyses methods for the resolution some problems of computer sciences are discussed.
6. The six rules of creation universal theories, which are created on the basis of polymetric analysis, are represented.
7. Perspective of development the synthetic system methods in computer sciences is analyzed too.

## References

**1.** Trokhimchuck P. P. S. Beer centurial problem in cybernetics and methods of its resolution. In: *Advanced in computer science*, vol. 7, ch. 5. Ed. Mukesh Singla (AkiNik Publications, New Delhi, 2020), pp. 87-118.

2. Beer S. We and complexity of modern world. In: *Cybernetics today: problems and propositions*. Znaniye, Moscow: 1976: 7 – 32. (In Russian)

3. Trokhimchuck P. P. Theories of Everythings: Past, Present, Future. Lambert Academic Publishing, Saarbrukken, 2021, 260.

4. George F. H. Philosophical Foundations of Cybernetics. Abacus Press, London, 1976, 157.

5. George F. H. Foundations of cybernetics. Gordon and Breach Science Publishing, London, Paris, New York, 1977, 260

6. Kukhtenko A. I. Cybernetics and Fundamental Science. Naukova Dumka, Kiev, 1987. 144 ( In Russian)

7. Moiseev N. N. Development algorithms. Nauka, Moscow, 1987, 302 ( In Russian)

8. Trokhimchuck P. P. Polymetrical Analysis and Cybernetics. Artificial Intelligence. №3-4 (85-86), 2019: 51 – 59.

9. Leubniz G. W. Studies of Universal Calculus / Gottfried Wilhelm Leubniz. Works in four volumes, volume 3. Moscow: Mysl', 1984, 3: 533-537 (In Russian)

10. Russel B. Introduction to mathematical philosophy. Museum street, London, 1948, 208.

11. Ruzha I. Foundations of mathematics. Vyshcha shkola, Kyiv, 1981, 352 (In Russian)

12. Trokhimchuck P. P. Automaton theories. Retrospective and perspective. In: Advanced in computer science, vol. 3. AkiNik Publications, New Delhi, 2019: 39-69.

13. Trokhimchuck P. P. Polymetrical Analysis. History, Concepts, Applications. Lambert Academic Publishing, Saarbrukken, 2018, 280.

14. Trokhimchuck P. P. Mathematical foundations of the knowledge. Polymetrical doctrine. Vezha, Lutsk, 2014, 624 (In Ukrainian)

15. Whitehead A. N. Science and the modern World. Pelican Mentor Books, N.-Y., 1948, 224

16. Wiener N. Cybernetics or control and the communication in the animal and machine. Sovetskoye Radio, Moscow, 1968, 328 (In Russian)

17. Moiseev N. N. Man and the noosphere. Molodaya guardia, Moscow, 1990, 351 (In Russian)

18. Klini S. Introduction to metamathematics. Izd-vo inostr. lit.-ry, Moscow, 1957, 390 (In Russian)

19. Genesereth M. R., Nillson N. I. Logical foundations of artificial intelligence. Morgan Kauffman Publishers, Inc., Palo Alto, 1988, 406.

20. Nillson N. I. Artificial Intelligence. A new synthesis. Morgan Kauffman Publishers, Inc., San Francisco, 1998, 514.

21. Nillson N. I. The Quest for Artificial Intelligence: A History of Ideas and Achievements. Cambridge University Press, New York, 2010, 562.

22. Novikov D.A. Cybernetics 2.0. Advances in Systems Science and Application. 2016. 16(1): 1 – 18.

23. Russel S. J., Norwig P. Artificial Intelligence. A modern Approach. Pearson Education Limited, Harlow, 2021, 1166.

24. Kasabov N. K. Time-Space, Spiking Neural Networks and Brain-Inspired Artificial Intelligence. Springer-Verlag GmbH, Berlin, 2019, 738.

25. Dorofeev V. P.  System of Strong Hybrid Artificial Intelligence and Moiseev, Efremov and Legasov Principles. In: Actual Problems of Fundamental science-2021, Conference Proceedings IV, edited by P. P. Trokhimchuck  et al., Vezha-Print, Lutsk, 2021:  47 – 50 (In Russian)

26. Dorofeev V., Trokhimchuck P.  Hybrid Super Intelligence and Polymetrical Analysis. 2021, 9p., E-print Cornell university, USA  (http://arxiv.org/abs/2111.09762)

27. Dorofeev V., Lebedev A., Shakirov V., Dunin-Barkowski W. Super Intelligence to solve COVID-19 Problem. Advances in Neural Computation, Machine Learning, and Cognitive Research IV, 2020: 293-300.

28. Brillouin L.  Science and Information Theory.  Courier Corporation, New York, 2004, 392.

29. Shannon C. E. A mathematical Theory of communication. The Bell System Technical Journal. 1948, 27:  379–423, 623–656.

30. Nepeyvoda N. Programming Styles and Techniques. Internet University of Information Technologies, Moscow, 2005, 320 (In Russian)

31. Ershov A. P. Introduction to theoretical programming. – Moscow: Nauka, 1977. – 288 p. (In Russian)

32. Kasti J.  Big systems. Connectivity, complexity and catastrophe. Wiley @ Sons, New York, 1979, 216 (In Russian)

33. Samoilenko Yu. I. Problems and methods of physical cybernetics. Institute of Mathematics of the National Academy of Sciences of Ukraine, Kiev, 2006. 644 (In Russian)

34. Dictionary on cybernetics.  The main edition of the M. P. Bazhan USE, Kyiv, 1989, 752 (In Russian)

35. Lapa V. G. Mathematical Fundamentals of Cybernetics.  High School, Kyiv, 1974, 452 (In Russian)

36. Haken H. Information and self-organization. Mir,  Moscow, 1991, 240 (In Russian)

37. Gödel K.  On Formally Undecidable Propositions in Principia Mathematica and Related Systems I. Dower Publications, inc., New York, 1992, 68.

38. Knut D. The art of computer programming.  Mir, Moscow, 1976, vol.1, 735; 1977, vol.2,  724 1973, vol.3, 844 (In Russian)

39. de Broglie L.  Thermodynamics of isolated point (Hidden thermodynamics of particles). In: L. de Broglie. Collected papers, vol. 4.  Print-Atel'e, Moscow, 2014: 8 – 111 (In Russian)

40. Computer science/ Wilkipedia – the free encyclopedia. // http:/en.wikipedia.org/wiki/ Computer_ science

41. Computing/ Wilkipedia – the free encyclopedia. // http:/en.wikipedia. org / wiki/ Computing

42. Emelyanov V. S.  Pugwash movement. In: About science and civilizations. Memories, thoughts and reflections of a scientists, ed. I. G. Usachev. Mysl, Moscow, 1986: 32-38 (In Russian)